\documentclass[10pt,journal,compsoc]{IEEEtran}

\usepackage[nocompress]{cite}
\usepackage{amsmath,amssymb,amsfonts}
\usepackage{algorithm}
\usepackage[noend]{algpseudocode}
\usepackage{graphicx}
\usepackage{subfig}
\usepackage{xcolor}

\begin{document}

\title{E3NE: An End-to-End Framework for Accelerating Spiking Neural Networks with Emerging Neural Encoding on FPGAs}

\author{Daniel~Gerlinghoff,~
        Zhehui~Wang,~
        Xiaozhe~Gu,~
        Rick~Siow~Mong~Goh,~
        Tao~Luo%
\IEEEcompsocitemizethanks{
\IEEEcompsocthanksitem D.~Gerlinghoff, Z.~Wang, R.S.M.~Goh and T.~Luo are with the Institute of High Performance Computing, Agency for Science Technology and Research (A*STAR), Singapore
\IEEEcompsocthanksitem X.~Gu is with the Future Network of Intelligence Institute, Chinese University of Hong Kong, Shenzhen, China
\IEEEcompsocthanksitem Corresponding author: Tao~Luo (leto.luo@gmail.com)}}

\markboth{IEEE Transactions on Parallel and Distributed Systems}%
{}

\IEEEtitleabstractindextext{%
\begin{abstract}
Compiler frameworks are crucial for the widespread use of FPGA-based deep learning accelerators. They allow researchers and developers, who are not familiar with hardware engineering, to harness the performance attained by domain-specific logic. There exists a variety of frameworks for conventional artificial neural networks. However, not much research effort has been put into the creation of frameworks optimized for spiking neural networks (SNNs). This new generation of neural networks becomes increasingly interesting for the deployment of AI on edge devices, which have tight power and resource constraints. Our end-to-end framework E3NE automates the generation of efficient SNN inference logic for FPGAs. Based on a PyTorch model and user parameters, it applies various optimizations and assesses trade-offs inherent to spike-based accelerators. Multiple levels of parallelism and the use of an emerging neural encoding scheme result in an efficiency superior to previous SNN hardware implementations. For a similar model, E3NE uses less than 50\% of hardware resources and 20\% less power, while reducing the latency by an order of magnitude. Furthermore, scalability and generality allowed the deployment of the large-scale SNN models AlexNet and VGG.
\end{abstract}

\begin{IEEEkeywords}
Spiking Neural Network, Neuromorphic Computing, Neural Encoding, Compiler Framework, FPGA
\end{IEEEkeywords}}

\newcommand{\IEEEcopyright}{
\noindent\parbox{\textwidth}{%
    \vspace{\baselineskip}
    \textcopyright~2021 IEEE. Personal use of this material is permitted. Permission from IEEE must be obtained for all other uses, in any current or future media, including reprinting/republishing this material for advertising or promotional purposes, creating new collective works, for resale or redistribution to servers or lists, or reuse of any copyrighted component of this work in other works.}
}

\IEEEcopyright
\maketitle

\IEEEraisesectionheading{\section{Introduction}\label{sec: introduction}}
\IEEEPARstart{W}{ith} the growing interest in deploying machine learning models on embedded devices, the need for power-efficient algorithms with constrained computing resources arose. Spiking neural networks (SNNs) became a promising machine learning technique in edge-AI due to their high efficiency advantages over conventional artificial neural networks (ANNs). A human brain is estimated to consume a mere 20 Watts of power~\cite{drubach2000brain}, making it orders of magnitude more efficient than modern machine learning hardware. Tapping onto the optimizations carried out by evolution, SNNs mimic biological processes which were discovered in experiments on nervous systems of mammals~\cite{abeles1993spatiotemporal, bair1994reliable}. It was observed that neurons generate electrical action potentials or \textit{spikes} with a duration of around one millisecond each~\cite{gerstner2002spiking}. A certain sequence of spikes, a \textit{spike train}, is used to transfer information between neurons~\cite{borst1999information}.

To use the complex results of experimental observations for computational applications, models like the Hodgkin-Huxley model~\cite{hodgkin1952quantitative}, the Izhikevich model~\cite{izhikevich2003simple} or the leaky integrate-and-fire model~\cite{koch1998methods} described the neuron behavior mathematically. Furthermore, neural encoding schemes such as rate encoding~\cite{zhang2003other} or temporal encoding~\cite{theunissen1995temporal} are used to convert real numbers into binary spike trains. Due to their differences with conventional ANNs, SNNs were coined the \textit{third generation of neural networks}~\cite{maass1997networks}.

The characteristics of SNNs enable low-power applications. The binary encoding of information in the form of spike trains obviates the need for expensive multipliers in the data path. To fully exploit the advantages of SNNs, its deployment on dedicated hardware systems is necessary. Application-specific integrated circuits (ASICs) like TrueNorth~\cite{merolla2014million} and SpiNNaker~\cite{stromatias2015scalable} are examples of adder-based neuromorphic devices which employ non-von Neumann architectures. Intel's neuromorphic chips Loihi~\cite{davies2018loihi} and Loihi~2~\cite{intel2021taking} combine multiple general microprocessor cores with 128 neuron cores. Those core communicate among each other via spikes through a Network-on-Chip. User-programmable processes allow the implementation of both inference and online training. In addition, architectures for Field-Programmable Gate Arrays (FPGAs) have been proposed for accurate emulation of spiking neuron characteristics~\cite{moore2012bluehive, luo2018fpga}. FPGAs are known for their malleability, which can be beneficial in this emerging research field, where new discoveries are made frequently. FPGAs can also easily be adapted to a wide range of applications, depending the requirements regarding performance, power consumption, cost, etc. However, the development for FPGAs requires special knowledge on hardware architecture and device properties. While high-level synthesis is able to derive a hardware structure from software code, it still requires directives and might not support all the features one expects from a software compiler. Moreover, it also comes at the expense of performance drop since the automatically generated hardware might not be fully optimized for the application. Therefore, it is challenging to make efficient FPGA-based SNN accelerators accessible to scientists and developers in the fields of neuroscience and artificial intelligence.

In the early days of spiking neural networks, the fundamental principles where tested on very simple problems, such as binary classification~\cite{gutig2006tempotron, lee2018deep}. As SNN algorithms matured, larger models like VGG were used to tackle complex datasets such as CIFAR-10 or ImageNet~\cite{sengupta2019going}. Hardware accelerators, however, seem to lag behind software simulations, as past results were mostly reported on toy datasets such as MNIST. A potential reason for this scalability hurdle is the spike train length, which can reach into the thousands if the network model is large. On the one hand, that is necessary to avoid information loss and attain a high classification accuracy. On the other hand, it increases the runtime of the network and hence offsets the efficiency, making it less attractive to deploy on SNN hardware. Emerging neural encoding schemes found ways to compress the spike trains where information are represented inefficiently. This opens possibilities for shorter spike trains with greater information density. Hence accuracy can be retained at a high value.

In this paper, we present an \textbf{E}nd-to-\textbf{E}nd framework for accelerating spiking neural networks with \textbf{E}merging \textbf{N}eural \textbf{E}ncoding on FPGAs. E\textsuperscript{3}NE enables the inference of SNN models to be deployed on almost arbitrarily-sized FPGAs. The framework builds upon efficient hand-crafted hardware blocks to execute the basic functions found in SNNs, including convolution and pooling. High performance is achieved by optimizing the instantiation and configuration of hardware blocks based on application requirements and hardware constraints. The framework employs a dynamic quantization scheme for the generation of spike trains and transformation of weights from a pre-trained SNN model. We provide open access to our GitHub repository\footnote{https://github.com/DanielGerlinghoff/radix-encoding}, which contains source files of both the hardware and software tools for E\textsuperscript{3}NE. Our contributions can be summarized as follows:
\begin{itemize}
    \item The first end-to-end framework for spiking neural networks that is based on register-transfer level (RTL) hardware blocks.
    \item Optimization of data movement and logic utilization by maximizing the degree of parallelism in the generated hardware.
    \item Support for emerging neural encoding to increase inference throughput and accuracy.
    \item Outperforming of previous SNN hardware implementations with regard to accuracy, latency, power, and hardware resources.
\end{itemize}

The rest of the paper is organized as follows. Section~\ref{sec: related work} discusses existing compiler frameworks for FPGA-based machine learning hardware and reviews past implementations of SNN inference hardware. Section~\ref{sec: background} introduces the basics of SNNs and neural encoding. We then describe the E\textsuperscript{3}NE framework in Section~\ref{sec: framework}, followed by the presentation of our experimental results in Section~\ref{sec: experiments}. Lastly, Section~\ref{sec: conclusion} concludes the paper.

\section{Related Work} \label{sec: related work}
Domain-specific design automation for neural network accelerators are crucial for the accessibility of fast FPGA hardware. Many previous works were presented to develop such compilers for traditional ANNs. Generally, they can be categorized into overlay-based solutions and dedicated hardware accelerators~\cite{plagwitz2021safari}.

The former employs universal hardware, which is able to efficiently execute the operations involved in convolutional and fully-connected layers of a network. The overlay is supplied with data and instructions and passes the computation result back to the host. Multiple network architectures can potentially be supported without the need for reconfiguration of the FPGA. Angel-Eye~\cite{guo2017angel}, DNNVM~\cite{xing2019dnnvm}, and DLA~\cite{abdelfattah2018dla} are examples of overlay-based compilers. All of them employ parameters to control the parallelism of feature maps, input/output channels, etc., based on the number of available processing units, which in turn is constrained by on-chip hardware resources. A compiler maps convolution layers to those processing units by determining tiling parameters and generating instructions for data movement and execution schedule. Specific custom instruction sets have been developed for their accelerators. TVM~\cite{chen2018tvm} is a framework for the deployment of neural networks across a variety of architectures, including GPU and FPGA. A general matrix-multiply, together with appropriate buffers and memory blocks, is instantiated and controlled by an instruction fetch unit.

Ma et al.~\cite{ma2017automatic}, on the other hand, use dedicated hardware blocks and configure them according to layer specifications. Their convolution modules have a fixed kernel size and stride. Those processing engines can be reused for layers with similar characteristics. Local configuration registers, which can be programmed during runtime, store information about the computation sequence. The proposed neural network compiler breaks down feature maps into tiles, which individually pass through a sophisticated sequence of diverse ANN operations.

While many design automation tools support conventional ANNs, there are only a few frameworks for their spiking counterparts. Due to the different computation paradigm, adapting an existing ANN framework would neglect important considerations about the time dimension. This has effects on the memory structure and access patterns. In addition, neural encoding, the update of neuron states, and spike train generation can be parts of the optimization process. Fang et al.~\cite{fang2020encoding} builds their hardware on top of a custom neural encoding scheme which utilizes the spike frequency as the information carrier. Their neuron model can hence be implemented as infinite impulse response filters, whose computation is efficiently implemented in digital signal processing (DSP) slices. Their compiler automatically determines bottlenecks in the computation flow and allocates more resources if the hardware platform permit. When dealing with pre-fabricated neuromorphic chips, the mapping of an SNN structure onto the fixed layout of neuron cores becomes another challenge. The LCompiler~\cite{lin2018mapping} is an SNN compiler for Loihi, which uses a greedy algorithm to iteratively optimize the mapping for power efficiency. With Lava~\cite{intel2021taking}, Intel recently announced a framework adopted for Loihi~2 and other neuromorphic platforms. Both BrainScaleS and SpiNNaker also come with their respective mapping algorithms~\cite{ehrlich2010software, galluppi2012hierachical}. The previous examples show the advantage of neuromorphic systems on FPGA, where hardware is reconfigured to fit the network architecture, instead of vice versa.

In addition to the automated compiling tools, there are hardware implementations that take advantage of the distinct properties of SNN. Minitaur~\cite{neil2014minitaur} is an event-driven architecture, where every neuron is idle until a spike occurs at its input. Since spike trains are often sparse, this can lead to significant power efficiency improvement. Han et al.~\cite{han2020hardware} use a hybrid approach for updating neurons. They iterate through the positions in the spike train in a sequential manner. Spikes at the same position are handled by an event queue. This improves the accuracy and hardware resources compared to Minitaur. However, both implementations only support fully-connected layers. In the realm of convolutional SNN accelerators, Ju et al.~\cite{ju2020fpga} introduce a two-dimensional array of convolution units, which reduces data movement by reusing input feature maps for multiple output feature maps. The simultaneous computation of multiple kernels and output channels reduces their runtime. However, their hardware design occupies many resources, making it difficult to be deployed on devices with small footprints. S2N2~\cite{khodamoradi2021s2n2}, on the other hand, is a SIMD architecture with a high resource efficiency. The architecture was also evaluated on a two-dimensional image dataset. Their work was implemented using high-level synthesis and the existing FINN framework for the acceleration of binary neural networks~\cite{umuroglu2017finn}. We will use existing SNN hardware implementations as a performance reference in the experiment section. Details on the low-level design of the hardware blocks, however, is not included in the scope of this paper.

\section{Background} \label{sec: background}
Spiking neural networks (SNN) are structurally similar to conventional neural networks. They consist of a sequence of layers through which information propagate. Each layer is either an array of neurons for fully-connected layers or a two-dimensional neuron feature map for convolution/pooling layers. At the neuron level, however, SNNs show inherently different characteristics from ANNs to achieve a greater resemblance of the biological processes found in human brains.

The primary method of communicating information between adjacent neurons in the spiking neural network is via \textit{spike trains} instead of real values in traditional ANNs. Those are sequences $\{s_0, s_1, ..., s_{T-2}, s_{T-1}\}$ of binary values $s_t$ to indicate the presence (one) or absence (zero) of a spike event. Their length is denoted as $T$, as in \text{time steps}. In contrast to ANNs, each SNN layer receives $T$ inputs in a sequence before being able to move on to process the next input sample. If depicted as a loop hierarchy, as often done for ANN implementations, time steps are added as an additional loop (see Algorithm~\ref{alg: loop hierarchy} in Section~\ref{sec: instructions}).

The generation of input spike trains from real values is an active research topic~\cite{gutig2014spike, sboev2020solving}. With \textit{temporal encoding}, the exact positions of spikes within the spike train relate to the represented value~\cite{sengupta2017spike, petro2019selection}. The commonly used \textit{rate encoding} sets the number of spike events over the length of the spike train proportional to the magnitude of the real value~\cite{diehl2015unsupervised, demin2018recurrent}. This process, however, leads to severe information loss if the spike train length $T$ is small, as only values in the range $[0, T]$ can be represented. Long spikes trains, on the other hand, require more time steps and hence increase the runtime. Previous publications used hundreds of time steps to maintain a high accuracy for large SNN models~\cite{sengupta2019going, rathi2020enabling}. Efficient neural encodings, such as~\cite{wang2021efficient}, are able to significantly shorten spike trains to a length of less then ten. The compression of information is achieved by scaling the impact of spikes on the neurons depending on their position in the spike train. When choosing those scaling factors to be powers of two, a spike train is transformed to $\{1 s_0,~ 2 s_1,~ ...,~ 2^{T-2} s_{T-2},~ 2^{T-1} s_{T-1}\}$. The impact of spikes on the neuron state doubles with every time step. That spike train is equivalent to the sequence of binary digits of a $T$-bit integer value, when read from least to most significant bit. That makes for a straightforward conversion between integers and efficient spike trains. The complexity is reduced to the selection of an appropriate method to quantize the real input values.

\section{End-to-End Design Framework} \label{sec: framework}
This section describes E\textsuperscript{3}NE in detail. First, we give an overview of the hardware library utilized by the framework. Then, we present the proposed compiler and corresponding optimization techniques. That covers the instantiation and configuration of processing modules, memory, and control instructions based on user input.

\subsection{Hardware Block Library} \label{sec: hardware}

\begin{figure}[!t]
\centering
\includegraphics[width=0.40\textwidth]{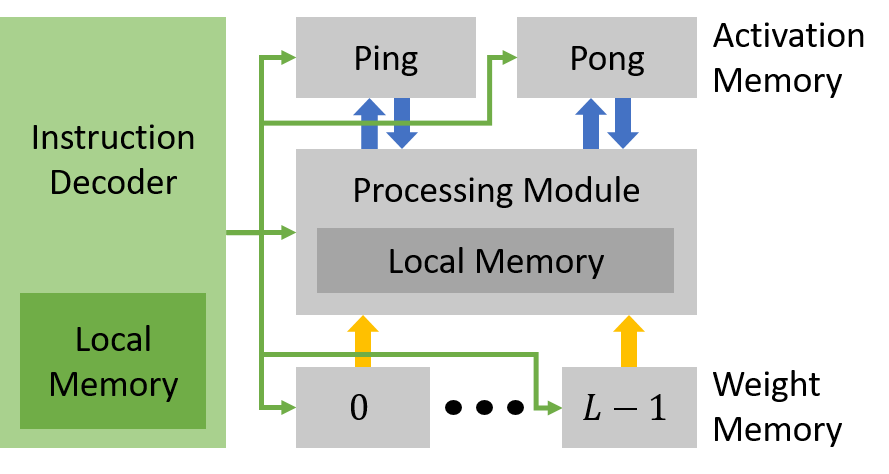}
\caption{Minimal example of the hardware used for E\textsuperscript{3}NE. Ping-pong buffers are used to bounce input and output activations of the respective layer processed in the processing module. Weights of $L$ layers are stored in $L$ separate memory blocks. Execution is controlled globally by an instruction decoder.}
\label{fig: hardware blocks}
\end{figure}

The propagation of information in the form of binary spike trains comes with a significant advantage for hardware implementations. While ANNs rely on resource-expensive multiply-accumulate operations to determine the output value of a neuron, SNN can instead work with conditional accumulate operations. The arrival of a spike at any time step triggers the addition of a $B$-bit weight value to the current time step's neuron state. In the absence of a spike, the neuron state remains unchanged. This enables us to employ arithmetic hardware blocks with a high energy and resource efficiency.

All hardware blocks are configured to adapt to the characteristics of the layers they compute. Thus, E\textsuperscript{3}NE falls into the category of \textit{dedicated hardware accelerators}. Processing modules for convolution and pooling can, however, be partially configured at runtime, allowing them to be reused for multiple network layers. Two-dimensional processing modules are characterized by their number of rows $Y$ and number of columns $X$. The dimensions are fixed at the time of compilation, as is the supported kernel size, which is set to the dimension $Y$. At the same time, two degrees of freedom are provided by the hardware to allow various layers with an equal kernel size to be executed on the same processing modules. Firstly, the kernel stride can be set dynamically, provided that all desired stride values have been implemented in the hardware during compilation. Secondly, feature maps of different sizes can be computed by the same module. The exact calculation of all static and dynamic configuration parameters are covered in Section~\ref{sec: processing modules}.

Our hardware design aims for extensive reuse of all hardware blocks, including both processing and memory modules. This resource efficiency leads to scalability in both directions, targeting both big and small FPGA devices. The data flow reflects this by sharing hardware among multiple layers. Figure~\ref{fig: hardware blocks} depicts a minimal implementation of our accelerator. Activation data is initially written into an activation memory block \textit{ping}. To compute a network layer, they are loaded into the processing module (blue arrows). That module can be any neural network operation, such as convolution, pooling and matrix-multiply. Weights are fetched as needed from a relevant weight memory block (yellow arrow). This example accommodates $L$ similar layers, each having a separate weight memory. Partial sums generated during processing are stored with high precision in the memory local to the processing module. The final output is written back to the \textit{pong} block. The next layer can be executed without any additional hardware resources by reusing the processing module with a different set of weights, letting activations bounce between buffers \textit{ping} and \textit{pong}. Careful dimensioning of processing and memory modules by the compiler ensures that feature maps of varying size fit while maximizing hardware utilization.

An instruction decoder issues commands to memory and processing modules based on instruction words, which are stored in the local memory. Instructions are used to control the data flow and launch layer operations. Furthermore, they set configuration registers prior to starting a computation to adapt the modules to changes in the layer configuration. The sequence of instructions is unique to the network architecture and is generated by the compiler upfront.

The aforementioned principles hold true when scaling the hardware to support larger networks with a more complex architecture. With more types of layers in the network, additional specialized processing modules are necessary. A set of activation ping-pong buffers is added if both one- and two-dimensional features are handled, which is usually the case in convolutional neural networks. The instruction decoder remains the central controller for all hardware blocks.

\subsection{Compilation Flow}

\begin{figure}[!t]
\centering
\includegraphics[width=\columnwidth]{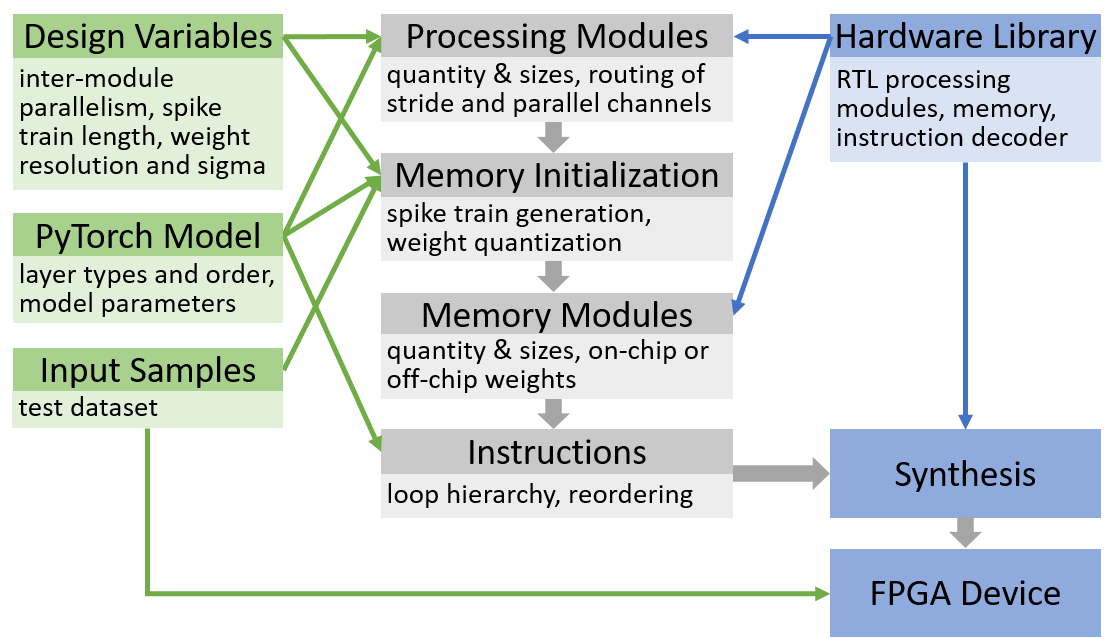}
\caption{Compilation flow showing the user input (green), hardware tools (blue), and compiler modules (gray). Arrows indicate the dependencies between modules.}
\label{fig: compile flow}
\end{figure}

The compiler's task is to assemble the building blocks in the hardware library based on the user specification, such that the machine learning task can be efficiently executed. An overview of the compilation process is given in Figure~\ref{fig: compile flow}. Compiler modules are shown in gray in the middle of this figure. We split the compilation into four sub-processes, each responsible to generate a certain part of the hardware architecture. Those sub-processes build upon each other, meaning that the sequence of execution was deliberately chosen. For example, instructions could not be generated without prior knowledge of processing and memory modules.

The user input is illustrated as green blocks in Figure~\ref{fig: compile flow} with green arrows indicating dependencies with compiler sub-processes. The starting point of the compilation is a neural network model, which was trained in the popular machine learning framework PyTorch~\cite{paszke2019pytorch}. The PyTorch model includes all hyper-parameters of the network, such as the architecture and layer specifications. It also contains the values of trained parameters, i.e. convolution kernels and weights for fully-connected layers. Design variables are used to constrain the hardware. Trade-offs between speed, accuracy, power and hardware resources can be achieved by adjusting variables such as resolution and parallelism. Input samples are usually stored in floating point format. The memory initialization sub-process converts them into spike trains so that they can be handled by the SNN accelerator.

Other than that, the hardware tools in blue are part of the compiler tool chain. The hardware library described earlier in Section~\ref{sec: hardware} contains the available hardware blocks in the format of HDL files. The synthesis tool takes those together with configuration files obtained from the compiler sub-processes to generate a bitstream for the FPGA device. In the following sections, we will describe every compilation step and corresponding optimizations to achieve a high degree of parallelism and efficiency.

\subsubsection{Processing Modules} \label{sec: processing modules}

\begin{figure}[!t]
\centering
\includegraphics[width=\columnwidth]{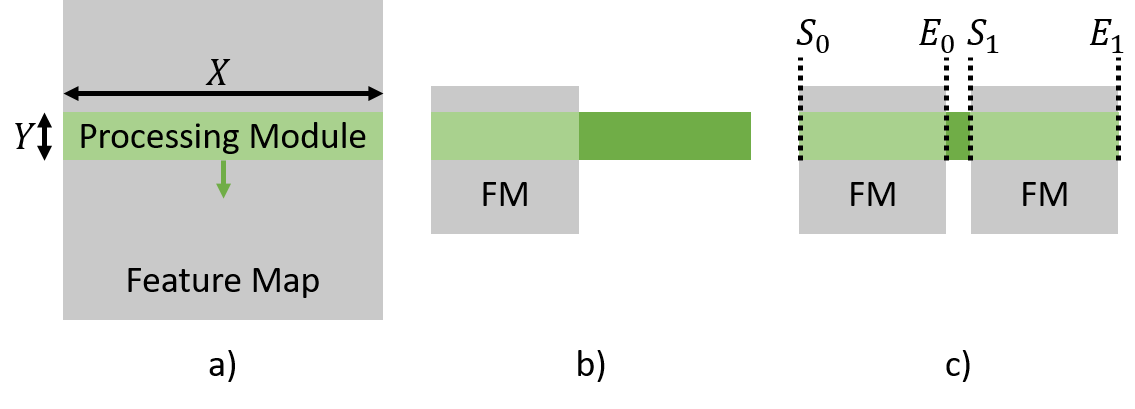}
\caption{Two-dimensional row-by-row operation on a feature map (gray) by a processing module (green) with height $Y$ and width $X$. a) Full utilization by matching $X$ to the size of the feature map. b) Naive and inefficient use of the same processing module on a much smaller feature map with unused hardware in darker green. c) Placing multiple output channels between start indices $S$ and end indices $E$ along the $X$ dimension of the processing module to increase utilization.}
\label{fig: processing module}
\end{figure}

The first step in the compilation process is to analyze the structure of the spiking neural network and derive the necessary processing modules (PMs) and their configurations. Each type of layer, i.e., convolution, pooling, or fully-connected layer, is computed with a dedicated PM. Through parallelism within modules, we optimize the hardware utilization for feature maps of different sizes. Parallelism between modules adjusts the computational performance. The outputs of this sub-process are configuration files with parameters, such as the number and sizes of processing modules and routing information for different kernel strides and parallelism.

\textbf{Configuration and Intra-Module Parallelism:}
Hardware blocks for two-dimensional operations, namely, convolution and pooling, are structurally alike and are defined by width $X$ and height $Y$. The latter is equal to the kernel size of the layer. With those dimensions being fixed in the hardware, at least one convolution/pooling module is instantiated for every kernel size occurring in the network. Algorithm~\ref{alg: processing module} describes this in lines 1 and 2 by iterating over all kernel sizes in the network and instantiating the respective PMs.

Due to the reusability of hardware blocks, multiple layers can share processing modules, as long as their kernel sizes match. To avoid the need for tiling of feature maps, the width $X$ is set to the size of the largest feature map processed by the module. In Algorithm~\ref{alg: processing module}, $D_{out}^{(l)}$ stands for the dimension of an output channel in layer $l$. The first layer with a kernel size $K$ determines $X$, as expressed in lines 4 and 5. Figure~\ref{fig: processing module} illustrates, how a processing module (green) moves across a feature map (gray) row-by-row starting at the first row. In case a), the size of the feature map matches the module and the hardware is fully utilized. 

Due to pooling operations and convolution with a kernel stride greater than one, feature maps tend to vary greatly between layers. This potentially leaves a large portion of the hardware resources idle, as depicted in Figure~\ref{fig: processing module} b) by the dark area in the PM. We alleviate this and maximize the logic utilization by computing multiple output channels simultaneously based on the same input channel. The computation flow is duplicated for all feature maps, which are placed side-by-side along the $X$ direction of the PM, as can be seen in Figure~\ref{fig: processing module} c). The position of each parallel feature map in the PM is determined by a start index $S_p$ and end index $E_p$. Algorithm~\ref{alg: processing module} formulates the calculation of those indices in lines 6 to 9. The number of parallel output channels $P^{(l)}$ of a layer $l$ depends on the widths of the PM and output channels. The calculation of indices $S_p$ and $E_p$ is influenced by the input channel dimension $D_{in}$, kernel stride $str$ and padding $pad$ of the respective layer $l$.

\begin{algorithm}
\linespread{1.25}\selectfont
\caption{Generation of 2D Processing Modules}
\label{alg: processing module}
\algnewcommand\algorithmicto{\textbf{to}}
\begin{algorithmic}[1]
    \ForAll{kernel sizes $K$ in the network}
        \State instantiate processing module with $Y \gets K$
        \ForAll{layers $l$ with kernel size $K$}
            \If{first iteration}
                \State $X \gets D_{out}^{(l)}$
            \EndIf
            \State parallel channels $P^{(l)} \gets \lfloor X / D_{out}^{(l)} \rfloor$
            \For{$p \gets 1~ \algorithmicto\ P$} \Comment{index $^{(l)}$ in loop omitted}
                \State $S_p \gets \lfloor p * (D_{in} + pad) / str \rfloor$
                \State $E_p \gets S_p + D_{out} - 1$
            \EndFor
            \State create routing for $P^{(l)}$ parallel output channels
            \State create routing for stride ${str}^{(l)}$
        \EndFor
    \EndFor
\end{algorithmic}
\end{algorithm}

Each two-dimensional PM has potentially multiple settings for parallelism, as well as kernel stride. To allow the data flow to be adjusted dynamically during runtime, all settings have to be reflected in the hardware at the time of synthesis. Therefore, the compiler incorporates the necessary routing beforehand while configuring the hardware blocks (see Algorithm~\ref{alg: processing module}, line 10 \& 11).

The one-dimensional processing module for fully-connected layers exhibits far less complexity compared to two-dimensional PMs. Configuration is limited to the number of parallel-computed output features. Since every operation requires a separate weight, the parallelism is bounded by the bandwidth of the memory used to store the weight values.

\textbf{Inter-Module Parallelism:}
Depending on the use case of our SNN accelerator, different constraints for performance, power, and hardware resources exist. E\textsuperscript{3}NE aims for high scalability to allow power and area efficient processing modules deployed on small and cost-effective FPGA devices. When higher performance is required, the inter-module parallelism can be realized by instantiating more than one processing module for each kernel size. The data flow is duplicated for all parallel PMs such that the output channels are computed concurrently. That is equivalent to unrolling the output channel loop as can be seen in Algorithm~\ref{alg: loop hierarchy}, line~3.

Through the design variables, the user can specify the number of parallel processing modules. In practice, however, only the duplication of convolution modules has a notable impact on the performance. For other PMs, the time to transfer data quickly exceeds the time spent on the actual computation. In this case, more memory bandwidth is necessary to achieve any further speed-up. The same limits apply to the convolution modules. However, due to the large amounts of computations involved in the convolution operation, this effect becomes obvious only with a larger number of parallel modules.

\subsubsection{Memory and Initialization}
The next two sub-processes of the compilation flow are related to the memory for activations and network parameters. Memory blocks are characterized by their width $W$ and height $H$. Our compiler sizes every memory module individually to minimize their resource consumption, while ensuring the fit of the data to be stored. Therefore, the initialization sub-process is executed first. That also includes weight quantization and spike train generation as part of the neural encoding scheme. Important trade-offs in terms of accuracy, latency, and memory resources are made during that process.

\textbf{Weight Memory:}
The PyTorch model holds the trained parameters of the convolution and fully-connected layers in floating point format. Quantization is needed before the weights can be used with our SNN accelerator. To minimize the information loss, E\textsuperscript{3}NE employs a dynamic quantization scheme based on~\cite{jacob2018quantization}. Specified by the user as a design variable, $B$ is the number of bits for the representation of weight and kernel values. This number is constant throughout the computation of the whole network. The range of weight values, however, might vary between layers. To pack as many information as possible into the $B$ available bits, floating point values $v_{wgt}$ are scaled in accordance with $R^{(l)}_{wgt}$ before being quantized. Equation~\ref{eqn: quantization weight} describes the full quantization rule for a weight value $v_{wgt}$ in layer $l$. Thereby, $\lfloor \bullet \rceil$ denotes rounding to the nearest integer and $(\bullet)^{\top \text{max}}_{\bot \text{min}}$ clamps a value between a minimum and maximum.

\begin{equation} \label{eqn: quantization weight}
    \tilde{v}_{wgt} = \left\lfloor v_{wgt} * 2 ^ {R^{(l)}_{wgt}} \right\rceil ^ {\top~2 ^ {B - 1} - 1} _ {\bot~-2 ^ {B - 1}}
\end{equation}

The scaling factor $R^{(l)}_{wgt}$ gives the layer-specific position of the radix point, i.e., the number of fractional binary digits, in quantized weights $\tilde{v}_{wgt}$. Its value is chosen such that, after scaling, a \textit{high percentage} of weight values in layer $l$ can be represented with $B$ bits, without the need for clamping. If we assume the weight values to be normally distributed, $R^{(l)}_{wgt}$ can be computed by Equation~\ref{eqn: scale weight}, with $\bar{v}^{(l)}_{wgt}$ and $\sigma^{(l)}_{wgt}$ being the mean and standard deviation of the weight set of layer $l$, respectively. The parameter $r$ can be trimmed by the user. Large $r$ avoids clamping but might reduce the number of fractional digits for just a few outliers. Smaller values might distort the model due to excessive clamping. A balance has to be found to maximize the inference accuracy.

\begin{equation} \label{eqn: scale weight}
    R^{(l)}_{wgt} = B - \left\lceil \log_2 \left(|\bar{v}^{(l)}_{wgt}| + r * \sigma^{(l)}_{wgt} \right) \right\rceil - 1
\end{equation}

Weights and kernels are stored in BRAM, from where they are supplied to the processing modules as needed. During the third sub-process of the compilation, memory modules are configured so that they can be instantiated in hardware. Depending on the number of network parameters and on the memory capacity, the compiler decides between two options for weight storage. If the memory capacity is sufficient, all network parameters will be stored in on-chip memory. In this case, each layer is assigned a dedicated read-only memory block, which is sized according to the number of number of parameters. In the case of convolution layers, the weights have the dimensions $(C_{out}, C_{in}, K, K)$, where $C$ is the number of channels and $K$ is the kernel size. One kernel of size $K * K$ is stored per row of ROM at $B$ bits per value. That leads to Equations~\ref{equ: weight memory} for the width $W$ and height $H$ of the memory block. In case of fully-connected layers, the width is determined according to the number of output features computed in parallel. That allows to read all values needed for the computation during a single clock cycle.

\begin{equation} \label{equ: weight memory}
\begin{split}
    W &= K ^ 2 * B \\
    H &= C_{out} * C_{in}
\end{split}
\end{equation}

The second option for weight storage is chosen, if not all parameters fit into on-chip memory. They are initially written to the external memory. Before the execution of every layer, respective weights are loaded dynamically into a single block RAM in the FPGA device, from where they can be accessed just like the ROM. For the memory dimensions, the maximum values for $W$ and $H$ in Equation~\ref{equ: weight memory} over all layers are chosen.

\textbf{Activation Buffers:}
Spike trains are encoded efficiently to reduce their length, as described in Section~\ref{sec: background}. Each position in the spike train corresponds to a binary digit of an $T$-bit integer number. Generating this $T$-bit integer follows the same dynamic quantization scheme as the weights. For every layer $l$, a scaling factor $R^{(l)}_{act}$ is computed with Equation~\ref{eqn: scale activation}. In contrast to the weights, which are fixed at the time of compilation, the magnitude of activations depends on the input given to the SNN model and is thus not strictly bounded. Instead, the maximum activation value $\hat{v}^{(l)}_{act}$ is computed for every layer $l$ across all input samples of a representative dataset. 

\begin{equation} \label{eqn: scale activation}
    R^{(l)}_{act} = T - \left\lceil \log_2 \hat{v}^{(l)}_{act} \right\rceil
\end{equation}

This ensures hat typical network inputs can be represented by a spike train of length $T$ without the need for clamping after being scaled with $R^{(l)}_{act}$. Equation~\ref{eqn: quantization activation} expresses the generation of a spike train $S$ based on an input value $v_{act}^{(1)}$ and the scaling factor $R^{(1)}_{act}$ of the first layer. The operation $(\bullet)_2$ denotes the representation as binary digits.

\begin{equation} \label{eqn: quantization activation}
    S = \left\lfloor v_{act}^{(1)} * (2 ^ {R^{(1)}_{act}} - 1) \right\rceil_2
\end{equation}

The accumulation of integers require the result to be represented with a higher number of bits than the operands in order to avoid information loss. To prevent the bit width of activations to in the course of evaluating a model, the activations are requantized after every convolution or fully-connected layer to rectify the position of the radix point for the next layer. Partial sum value $v_{psum}^{(l)}$ is the high-precision result of the accumulation. Precision is reduced by a right-shift operation in Equation~\ref{eqn: requantization}, such that the position of the radix point matches the activation scaling factor of layer $l+1$. Trimming digits leads to numbers being rounded down, increasing the requantization error. Rounding to the nearest integer is achieved by a conditional add operation.

\begin{equation} \label{eqn: requantization}
    v_{act}^{(l+1)} = \left\lfloor v_{psum}^{(l)} \gg \left(R^{(l)}_{wgt} + R^{(l)}_{act} - R^{(l+1)}_{act}\right) \right\rceil
\end{equation}

Spike trains are stored in alternating buffers after being requantized. The sizes of those buffers are determined by simulating the network execution and recording the size required to store activations, as done in Algorithm~\ref{alg: activation buffers} for two-dimensional feature maps. The buffer is expanded to at least the size required to store the respective feature maps. One row of binary spike activations is stored per memory address, requiring width $W$ to be at least of dimension $D_{in}$. The height $H$ depends additionally on the number of channels $C_{in}$ and time steps $T$. The target buffer changes between \textit{ping} and \textit{pong} after each layer.

\begin{algorithm}
\linespread{1.25}\selectfont
\caption{Generation of 2D Ping-Pong Buffers}\label{alg: activation buffers}
\begin{algorithmic}[1]
    \State ($W_{ping}, H_{ping}) \gets (0, 0)$
    \State ($W_{pong}, H_{pong}) \gets (0, 0)$
    \State $pp \gets ping$
    \ForAll{layers $l$ with dimension $D_{in}^{(l)}$ and channels $C_{in}^{(l)}$}
        \If{$pp = ping$}
            \State $W_{ping} \gets \max(W_{ping}, D_{in}^{(l)})$
            \State $H_{ping} \gets \max(H_{ping}, D_{in}^{(l)} * C_{in}^{(l)} * T)$
            \State $pp \gets pong$
        \ElsIf{$pp = pong$}
            \State $W_{pong} \gets \max(W_{pong}, D_{in}^{(l)})$
            \State $H_{pong} \gets \max(H_{pong}, D_{in}^{(l)} * C_{in}^{(l)} * T)$
            \State $pp \gets ping$
        \EndIf
    \EndFor
\end{algorithmic}
\end{algorithm}

\subsubsection{Instructions} \label{sec: instructions}

\begin{figure}[!t]
\subfloat[The configuration instruction assigns the value to a parameter.]{%
    \includegraphics[width=\columnwidth]{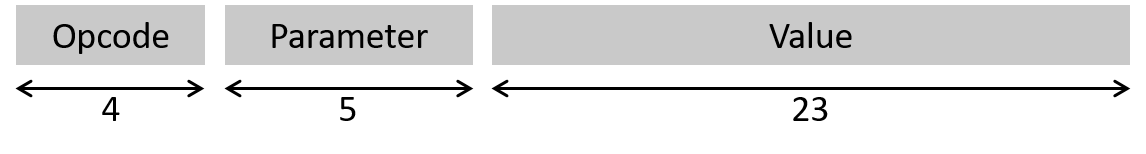}%
}

\subfloat[The command instruction starts and resets the processing in all modules.]{%
    \includegraphics[width=\columnwidth]{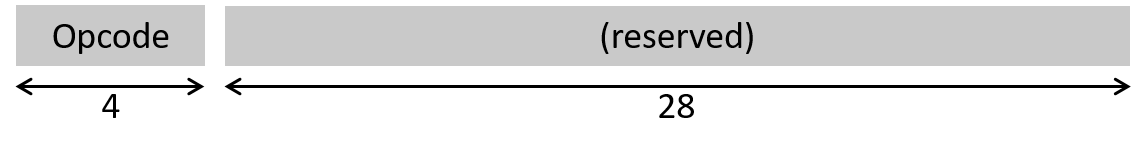}%
}

\subfloat[The memory instruction starts a transfer from or to a memory module at the address provided.]{%
    \includegraphics[width=\columnwidth]{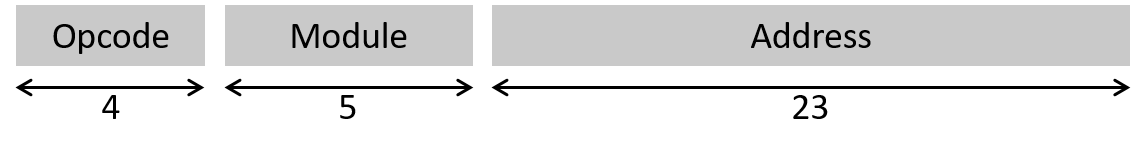}%
}

\subfloat[The wait instructions stalls the processor until the condition in the module becomes true.]{%
    \includegraphics[width=\columnwidth]{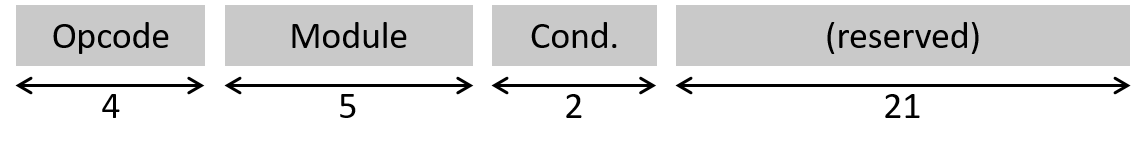}%
}

\caption{Formats of each instruction category used by the processor.}
\label{fig: instruction format}
\end{figure}

Every action of the accelerator hardware is controlled by instructions. That includes the dynamic configuration of the aforementioned hardware blocks, data movement and launch of neural network operations. The sequence of instructions is derived for every layer from its loop hierarchy. That is an expression of the nested iterations taken over channels, feature maps, kernels, etc. Our loop hierarchy for a convolution layer is shown in Algorithm~\ref{alg: loop hierarchy}. Convolution layers have the deepest loop hierarchy among all layer types, which is the reason for the high computational effort required. Optimization can be applied to reduce data movement and necessary memory capacity, as done before for ANNs by~\cite{zhang2015optimizing, li2018smartshuttle}. We interchanged loops such that accumulating (acc.) loops, like input channels and time steps, are executed before output channels, i.e., making them the inner loops. That reduces the memory footprint, since less high-precision partial sums have to be stored intermediately. Unrolling (unr.) is another loop optimization enabled by parallelism in hardware, as described earlier in Section~\ref{sec: processing modules}. Since not all output channels are computed in parallel, the loop is only partially unrolled.

\begin{algorithm}
\caption{Convolution Loop Hierarchy and Instructions} \label{alg: loop hierarchy}
\begin{algorithmic}[1]
    \State configure hardware
    \State select of memory modules
    \ForAll{output channels}\Comment{partially unrolled}
    \ForAll{time steps}\Comment{accumulating}
    \ForAll{input channels}\Comment{accumulating}
    \State reset
    \State load kernels into PMs
    \State load first activation row from \textit{ping}
    \ForAll{activation rows}
    \State start processing
    \ForAll{kernel rows}\Comment{acc. \& unr.}
    \ForAll{kernel cols}\Comment{acc.}
    \ForAll{activation cols}\Comment{unr.}
    \State accumulate over the kernel
    \EndFor
    \EndFor
    \EndFor
    \State load next activation row from \textit{ping}
    \State wait for processing to finish
    \State write partial sum
    \EndFor
    \State accumulate with previous input channels
    \EndFor
    \EndFor
    \State requantize activations
    \State write activations to buffer \textit{pong}
    \EndFor
\end{algorithmic}
\end{algorithm}

The compiler follows the loop hierarchies when analyzing the network layer-by-layer and generates 32-bit instructions for every operation. Each instruction belongs to one of the four categories shown below. Instructions of different categories differ with respect to their function and format. Apart from the operation code, the available bits are used to indicate configuration parameters, wait conditions, etc. Figure~\ref{fig: instruction format} visualizes the anatomy of each instruction category. The full instruction set is described in Table~\ref{tab: isa} with the last column indicating the category of the instruction.
\begin{itemize}
    \item Configuration of processing modules and memory
    \item Commands to launch computations and reset
    \item Memory operations for kernels and activations
    \item Wait for processes to complete
\end{itemize}

\begin{table}[!t]
    \renewcommand{\arraystretch}{1.25}
    \caption{Description of the Full Instruction Set.}
    \label{tab: isa}
    \centering
    \setlength{\tabcolsep}{5pt}
    \begin{tabular}{l|l l}
        \hline
        \textbf{Instructions} & \textbf{Description} & \textbf{Category} \\
        \hline
        \textbf{ENA}  enable,          & Set configuration parameters & Config. \\[-1ex]
        \textbf{CONF} configure        & in processing modules                  \\
        \textbf{PROC} process          & Start processing of convol.  & Cmd.    \\[-1ex]
        \textbf{LIN}  linear           & and linear layers                      \\
        \textbf{RST}  reset            & Reset processing modules     & Cmd.    \\[-1ex]
        \textbf{END}  end              & and last layer reached                 \\
        \textbf{KERL} kernel-load      & Load kernels from BRAM       & Mem.    \\[-1ex]
        \textbf{KERD} kernel-dram      & and DRAM                               \\
        \textbf{ACTL} activation-load, & Load and store activations   & Mem.    \\[-1ex]
        \textbf{ACTS} activation-store & from/to BRAM                           \\
        \textbf{WAIT} wait             & Wait for processes to finish & Wait    \\
        \hline
    \end{tabular}
\end{table}

At the beginning of each convolution layer, the hardware is configured to the specifications of the layer. That includes setting the parallelism, stride, and scaling factors for requantization. Subsequently, it is selected from which memory to fetch the weights and which memory modules act as ping-pong buffers. For every input channel, the processing modules (PMs) involved in the current layer's computation are reset and receive their respective kernel values. The processing is started separately for each activation row, after the data have been loaded from the buffer \textit{ping}. An opportunity for parallel execution arises, when the processor stalls while the computation is ongoing. During that waiting time, the next activation row is loaded. This engagement of multiple hardware circuits at the same time can shorten the runtime. Results of the convolution operation are written into a partial sum memory, where they are accumulated over all input channels. After all time steps have been processed, instructions are generated which requantize the output feature maps and write them back to buffer \textit{pong}.

To avoid data hazards, our instruction decoder does not have multi-processing capabilities. We implemented a scalar processor without support for pipelining. Instructions in the processor-internal RAM are fetched one-by-one. In a typical neural network application, one can expect roughly 0.4 instructions per clock.

\section{Experiments} \label{sec: experiments}
In this section, we explore how different settings of the compilation parameters affect the performance and efficiency of the spiking neural network inference. The experiments in Sections~\ref{sec: experiment intra parallelism} to~\ref{sec: experiment spike train length} are carried out on the MNIST dataset~\cite{lecun1998mnist} using the LeNet-5 network~\cite{lecun1998gradient}. MNIST consists of 60,000 training images and 10,000 test images of handwritten digits. It is a common dataset to verify SNN-related algorithms and hardware. LeNet is an early convolutional neural network designed to classify images of the MNIST dataset and has the architecture 32×32×1 -- 6C5 -- P2 -- 16C5 -- P2 -- 120C5 -- 120 -- 84 -- 10. In this notation, $c\text{C}k$ is a convolution layer with $c$ output channels and a kernel size of $k$. $\text{P}k$ denotes a pooling layer with both kernel size and stride equal to $k$. How well E\textsuperscript{3}NE can cope with large SNN models is evaluated on VGG~\cite{simonyan2014very} and AlexNet~\cite{krizhevsky2012imagenet}, using the CIFAR-100 and CIFAR-10 dataset, respectively. We also compare our framework with previous SNN hardware accelerators by deploying a very similar convolutional SNN for a fair comparison.

The SNN accelerator is deployed on a Xilinx Virtex UltraScale+ XCVU13P FPGA. Since multipliers are not needed for processing SNN models, no DSP slices have been used to implement the arithmetic functions. The Vivado tool chain was used to synthesize and implement the configured hardware blocks using the default settings. A clock frequency of 200 MHz is used for the FPGA platform, unless otherwise stated.

\subsection{Intra-Module Parallelism} \label{sec: experiment intra parallelism}
LeNet has a homogeneous network architecture, because all convolution layers use a kernel size $K = 5$ and for all pooling layers $K = 2$. For that reason, the LeNet can be evaluated using only two processing modules (PMs) with $Y = 5$ and $Y = 2$ for convolution and pooling, respectively. The width $X$ of the PMs is set according to size of the largest output feature map that they process. In that case, $X = 31$ for the convolution module and $X = 14$ for the pooling module. With 3-bit weights and a spike train length of $T = 4$, our LeNet SNN can achieve a reasonable accuracy of 99.09\%. The weight quantization was carried out with the parameter $r = 3$ (see Equation~\ref{eqn: scale weight}), i.e., values in the range of three standard deviations can be represented by three bits without clamping.

By default, the compiler uses intra-module parallelism to increase the hardware utilization, as described in Section~\ref{sec: processing modules}. In the case of LeNet, the convolution module is reused by all three convolution layers due to their equal kernel size. More than one output channel can be computed simultaneously for the second and third convolution layer. The output channel sizes of all two-dimensional layers are listed in Table~\ref{tab: experiment intra parallelism}. With the widths of the PMs fixed, the compiler determined the settings for intra-module parallelism as seen in the third column of that table. The smaller the output channel size, the more parallel channels can be placed in the PM. This does not affect the classification results and the accuracy, since the operations stay the same.

This experiment demonstrates the effect of intra-module parallelism. In Table~\ref{tab: experiment intra parallelism}, we compare the hardware utilization and runtime of each layer with and without parallel computing. In the second case, the PMs only compute one output channel at a time, as shown before in Figure~\ref{fig: processing module} b). This negatively affects the hardware utilization of the last three layers. Even with parallelism, however, the placing along the width of the PM depends on the input channel size and padding (see Algorithm~\ref{alg: processing module}, line 8). This limits a further increase in hardware utilization.

More pronounced is the effect on the runtime. The total runtime of LeNet increases fourfold with intra-module parallelism disabled. It can be observed that especially convolution layers benefit. The runtime is almost inversely proportional to the degree of parallelism. In case of the last convolution layer, for example, using one instead of six parallel channels increases the runtime by 5.2 times. Pooling layers, on the other hand, only see a slight speedup when using parallel computing. Because the number of operations is significantly less compared to convolution layers, the data transfer has a larger impact on the runtime. For the same reason, we will apply inter-module parallelism in the next section to convolution layers only.

\begin{table}[!t]
    \renewcommand{\arraystretch}{1.25}
    \caption{Intra-Module Parallelism (Intra-MP) of convolution and pooling layers in LeNet with an input image size of 32$\times$32 pixels.}
    \label{tab: experiment intra parallelism}
    \centering
    \setlength{\tabcolsep}{4.6pt}
    \begin{tabular}{l r|r r r|r r}
        \hline
        \textbf{} & \textbf{} & \multicolumn{3}{c|}{\textbf{With Intra-MP}} & \multicolumn{2}{c}{\textbf{Without Intra-MP}} \\
        \textbf{Layer} & \textbf{Size} & \textbf{Para.} & \textbf{Util. [\%]} & \textbf{Time [\textmu s]} & \textbf{Util. [\%]} & \textbf{Time [\textmu s]} \\
        \hline
        6C5   & 28 & 1 &  90 &  46 &  90 &   46 \\
        P2    & 14 & 1 & 100 &  13 & 100 &   13 \\
        16C5  & 10 & 2 &  65 & 154 &  32 &  301 \\
        P2    & 5  & 2 &  71 &  10 &  36 &   12 \\
        120C5 & 1  & 6 &  39 & 476 &   6 & 2464 \\
        \hline
        \multicolumn{2}{l|}{\textbf{Total}} & & 69 & 699 & 49 & 2836 \\
        \hline
    \end{tabular}
\end{table}

\subsection{Inter-Module Parallelism} \label{sec: experiment inter parallelism}

\begin{figure}[!t]
\centering
\includegraphics[width=\columnwidth]{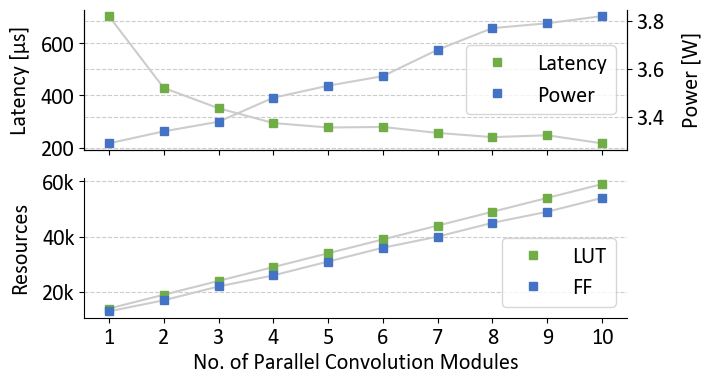}
\caption{Latency, power and hardware resources (lookup tables and flip-flops) when varying the number of parallel convolution modules from 1 through 10.}
\label{fig: plot parallelism}
\end{figure}

Through a design variable, the user can instruct the compiler to instantiate more than one convolution module for every kernel size. In contrast to intra-module parallelism, the objective is not to increase the utilization of existing hardware. Instead, additional hardware resources are required for inter-module parallelism. Those trade-offs are explored in this experiment. We use the same LeNet network architecture, compilation settings, and hardware platform as in the previous section. Intra-module parallelism is enabled by default.

Various metrics are plotted in Figure~\ref{fig: plot parallelism} with dependence on the number of parallel convolution modules. The main goal of instantiating multiple modules is achieved as can be seen from the latency plot. The latency tends to decrease when more convolution modules are used. Compared to the baseline of one module (705 \textmu s), the processing time is reduced to only 31\% when using ten parallel modules (216 \textmu s). One can observe a convergence of the latency values with a growing number of parallel modules. This is expected and in accordance with the Roofline model, which provides intuitive explanations for the limitations of hardware architectures~\cite{williams2009roofline}. In its essence, it assumes that hardware performance is limited either by the accelerator's peak performance or the memory bandwidth. Ideal architectures reach a performance close to one of those limits. Whether the peak performance or the memory bandwidth is the bottleneck depends on the Computation-to-Communication (C/C) ratio. The time used for computation of the convolution operations decreases when using more parallel modules. The time spent on communication is roughly constant, since the same amount of data has to be moved between processing and memory modules. Hence, the C/C ratio of our SNN accelerator decreases, which leads to the application being memory bounded. This effect is visualized by the latency values, which converge to the communication time of around 200 \textmu s. It can also be observed that the latency slightly increases between five \& six and eight \& nine parallel modules. In these cases, additional modules remain unutilized and lead to no reduction of the computation time. However, additional clock cycles are set aside to configure these additional hardware blocks, causing an increase in overall latency.

The plot of power values in Figure~\ref{fig: plot parallelism} shows the expected increase in power consumption with increasing number of instantiated hardware blocks. Interestingly, a larger increase in power goes together with a larger drop in latency and, vice versa, a smaller power increase coincides with no changes in latency. This correlation does not provide a definitive explanation to the cause and effect. However, we can reason that the dynamic power of additional modules is negligible when they are not utilized. Only the static power, which is consumed regardless of the module's activity, causes the minor increase in overall power.

Hardware resources shown in Figure~\ref{fig: plot parallelism} are measured in terms of the number of 6-input lookup tables (LUTs) and flip-flops (FFs). A very linear behavior can be observed, with around 4900 additional LUTs and 4500 flip-flops for every added convolution module. The remaining hardware resources are shared mainly between a pooling module (200 LUTs, 500 FFs), a linear module (4800 LUTs, 2900 FFs), and the instruction decoder (500 LUTs, 200 FFs). 

This experiment confirmed most of our expectations regarding the metrics for a variable number of parallel convolution modules. For high-performance applications, where hardware cost and power consumption are of less importance, a high degree of parallelism can be chosen. Embedded devices, which operate with a tight power budget, can execute the task with identical accuracy at the cost of a larger latency.

\subsection{Instruction Parallelism} \label{sec: experiment instruction parallelism}
A naive compiler implementation would generate instructions in sequence, following the loop hierarchy of the network layers. This experiment analyzes the impact of reordering instructions such that multiple hardware circuits are utilized at the same time. Network architecture and hardware configuration is identical to the previous two experiments. 

For every row of a two-dimensional feature map, a start instruction activates the processing module (see Algorithm~\ref{alg: loop hierarchy}, line 10). It needs no additional supervision by the instruction decoder until the activation row is fully processed. Hence, the processor stalls and waits for the computation to finish. In the case of LeNet, the waiting time without reordering equals eight clock cycles for convolution layers. The waiting time mainly depends on the number of kernel columns, whose loop is not unrolled and thus executed sequentially (see Algorithm~\ref{alg: loop hierarchy}, line 12). Our compiler makes the processor use those idle clock cycles for loading the next activation row. Since this process has a duration of two clock cycles, the idle time can be reduced to six cycles.

The influence of instruction parallelism on the latency varies depending on the number of convolution units, as can be observed from the measurements in Table~\ref{tab: experiment instruction parallelism}. In case of a single convolution module, instruction parallelism reduces the latency by 11\%. However, the effect decreases with more parallel convolution modules being deployed. Only 5\% of latency reduction are achieved with eight modules. A potential reason is the overall decrease in the number of start and, therefore, wait instructions executed by the processor when parallelism is increased. That means, the ratio between total waiting time and total latency decreases. Reducing the remaining waiting time by two clock cycles has a limited effect on the total latency. Put in terms of C/C ratio covered in the previous section, the improvement of computation time has a smaller effect when the communication time dominates the total runtime.

\begin{table}[!t]
    \renewcommand{\arraystretch}{1.25}
    \caption{Latency with and without instruction parallelism (IP) for various degree of inter-module parallelism \mbox{(Inter-MP)}.}
    \label{tab: experiment instruction parallelism}
    \centering
    \setlength{\tabcolsep}{7.5pt}
    \begin{tabular}{c|c|c c}
        \hline
        \textbf{} & \multicolumn{1}{c|}{\textbf{Without IP}} & \multicolumn{2}{c}{\textbf{With IP}} \\
        \textbf{Inter-MP} & \textbf{Latency [\textmu s]} & \textbf{Latency [\textmu s]} & \textbf{Latency Reduction} \\
        \hline
        1 & 789 & 705 & 11\% \\
        2 & 471 & 429 &  9\% \\
        4 & 316 & 294 &  7\% \\
        8 & 252 & 240 &  5\% \\
        \hline
    \end{tabular}
\end{table}

\subsection{Spike Train Length} \label{sec: experiment spike train length}

\begin{figure}[!t]
\centering
\includegraphics[width=\columnwidth]{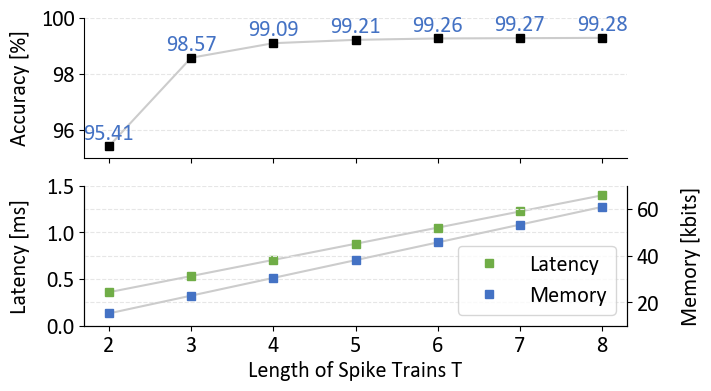}
\caption{Validation accuracy, latency and the size of activation memory dependent on the spike train length, i.e. number of time steps $T$.}
\label{fig: plot timesteps}
\end{figure}

This experiment explores the relationship between spike train length and classification accuracy, which is a common trade-off in spiking neural networks. The spike train length was used interchangeably with the number time steps $T$ in Section~\ref{sec: background}, which indicates an additional impact on the execution time. The LeNet model and hardware settings are reused from the previous experiments. No inter-module parallelism is applied here.

Figure~\ref{fig: plot timesteps} shows the results for a spike train length ranging from two to eight. At the lower end of this range, the validation accuracy drops sharply to only 95.4\%. This is a practical lower limit, since binary activation, i.e., $T = 1$, would lead to an unacceptable accuracy of less than 50\%. The classification of the MNIST test images becomes more accurate as the spike trains become longer. Because longer spike trains can represent a wider range of activation values, the quantization error is reduced. On the upper end of the length range, we achieve a maximum accuracy of 99.28\%. The curve flattens out towards the end and further lengthening of the spike trains would not lead to any accuracy gain due to the inherent limitations of the LeNet model.

Reaching the maximum accuracy with only eight time steps is made possible by efficient encoding. Because $2^T-1$ values can be represented per spike train, the error caused by quantization after every layer is reduced exponentially when adding more time steps. In comparison, Fang et al.~\cite{fang2020encoding} needed nine time steps to reach an accuracy of approximately 99.2\%. Our encoding attained the same accuracy with only five time steps, or around 55\% of their spike train length.

Longer spike trains, however, lead to a linear increase in latency as can be seen in the lower plot of Figure~\ref{fig: plot timesteps}. Time steps are processed in sequence by the hardware accelerator. Hence, more time passes until the classification result becomes available at the output. With the compilation settings used for this experiment, each additional time step leads to a latency increase of around 170 \textmu s. Moreover, spike trains are stored in ping-pong buffers between layers. Their size needs to be adjusted to account for the additional data. The lengthening of spike trains increase the capacity of activation memory by increments of 7.6 kbits.

\subsection{Performance Comparison and Scalability}

\begin{table*}[!t]
    \renewcommand{\arraystretch}{1.25}
    \caption{Comparison among SNN hardware implementations with regards to performance and efficiency.}
    \label{tab: comparision}
    \begin{center}
    \begin{tabular}{l l l c|r r r r c}
        \hline
        \textbf{Platform} & \textbf{Dataset} & \textbf{Network} & \textbf{Accuracy} & \textbf{Frequency} & \textbf{Latency} & \textbf{Throughput} & \textbf{Power} & \textbf{LUTs / FFs} \\
        {} & {} & {} & \textbf{[\%]} & \textbf{[MHz]} & \textbf{[\textmu s]} & \textbf{[fps]} & \textbf{[W]} & {} \\
        \hline
        Ju et al. \cite{ju2020fpga}         & MNIST     & CNN\textsuperscript{1} & 98.9 & 150  & 6110 & 164  & 4.6  & 107k / 67k  \\
        Fang et al. \cite{fang2020encoding} & MNIST     & CNN\textsuperscript{2} & 99.2 & 125  & 7530 & 2124 & 4.5  & 156k / 233k \\
        \textbf{E3NE} (this work)           & MNIST     & CNN\textsuperscript{2} & 99.3 & 200  &  409 & 2445 & 3.6  & 41k / 36k   \\
        \hline
        \textbf{E3NE}                       & MNIST     & LeNet-5                & 99.1 & 200  &  294 & 3400 & 3.4  & 27k / 24k   \\
        \textbf{E3NE}                       & CIFAR-10  & AlexNet                & 80.6 & 150  &  70k & 14.3 & 4.7  & 48k / 50k   \\
        \textbf{E3NE}                       & CIFAR-100 & VGG-11                 & 65.0 & 150  & 163k & 6.1  & 5.0  & 88k / 84k   \\
        \hline
        RTX 5000 \cite{fang2020encoding}    & MNIST     & CNN\textsuperscript{2} & 99.2 & 1620 & 18k  & 864  & 61.2 & (GPU)       \\
        i7-6700K \cite{ju2020fpga}          & MNIST     & CNN\textsuperscript{1} & 98.9 & 4000 & 252k & 4    & 54   & (CPU)       \\
        TrueNorth \cite{merolla2014million} & MNIST     & CNN & 99.4 & ---\textsuperscript{4}  & 1000 & 1000 & 0.2  & (ASIC)      \\
        Loihi \cite{shrestha2021hardware}   & MNIST     & CNN\textsuperscript{3} & 94.7 & 0.01 & 10k  & 97   & 0.24 & (ASIC)      \\
        \hline
    \end{tabular}
    \end{center}
    \hspace*{12mm}\textsuperscript{1} \footnotesize{28×28 -- 64C5 -- P2 -- 64C5 -- P2 -- 128 -- 10},
    \textsuperscript{2} \footnotesize{28×28 -- 32C3 -- P2 -- 32C3 -- P2 -- 256 -- 10},
    \textsuperscript{3} \footnotesize{32×32 -- 16C5 -- P2 -- 8C3 -- P2 -- 100 -- 10} \\
    \hspace*{12mm}\textsuperscript{4} \footnotesize{Asynchronous event-based execution without a global clock} \\
\end{table*}

The classification of the handwritten digits in the MNIST dataset was attempted before on SNN hardware. Table~\ref{tab: comparision} compares those research efforts with regards to common metrics, such as accuracy, latency, power and hardware resources. Furthermore, a cross-platform comparison is provided, which demonstrates the performance of SNN hardware in relation to other popular machine learning architectures.

Ju et al.~\cite{ju2020fpga} employs a custom convolutional neural network (CNN) with 169k parameters. We would expect their accuracy to be higher than the one of LeNet, whose number of parameters is less than half. However, their accuracy of 98.9\% is around 0.2\% lower. One potential reason for that could be their use of firing rate encoding, which is paired with a random number generator for the generation of Poisson-distributed spike trains. A statistical process like that tends to cause information loss if the spike train is as short as ten time steps. We mitigated this problem by using a more efficient neural encoding scheme. Fang et al.~\cite{fang2020encoding} designed a custom CNN architecture with 217k parameters. They can achieve 99.2\% of accuracy. To carry out a fair comparison, we deployed the same model on our SNN accelerator. We used eight parallel convolution modules and set $T = B = 4$ for the resolution of activations and weights, respectively. Not only did we increase the accuracy to 99.27\%, but we also outperformed them with regards to all other metrics. We reduced their latency by 94\% to only 409 \textmu s. With this latency, we are able to process 2445 images per second. That is 15$\times$ more than in~\cite{ju2020fpga}. At the same time, we only consume 80\% of the power reported in~\cite{fang2020encoding} and~\cite{ju2020fpga}. The former also uses almost 4$\times$ more LUTs and 6$\times$ more flip-flops (FFs), which could be a consequence of using high-level synthesis in their compilation flow. Overall, this comparison summarizes the advantages of E\textsuperscript{3}NE. Optimizations and parallelism are applied to almost all sub-processes of the framework. Combined with the use of an efficient neural encoding scheme, that yields a remarkable reduction in latency and power, while maintaining a high level of accuracy. The reuse of hardware blocks, which themselves have small area requirements, leads to a small footprint suitable for deployment on low-cost FPGAs.

Our comparison is limited to SNN hardware which is able to handle convolutional layers, since they are used without exception in modern neural network architectures. The classification of MNIST images is, however, possible with fully-connected layers only. Three-layer networks were implemented by Minitaur~\cite{neil2014minitaur} and Han et al.~\cite{han2020hardware}. Their power consumption benefits form the simple network structure and the reduced computational effort. But accuracy was negatively impacted such that their result is not on par with the listed convolutional SNN implementations.

The advantages of custom SNN hardware deployed on FPGA becomes obvious, when compared with other computing platforms as done in Table~\ref{tab: comparision}. Nvidia RTX 5000 is a GPU widely used for deep learning applications. When presumably used with a large batch size, it can process over 800 images per second, but latency and power consumption are at least an order of magnitude higher than for E\textsuperscript{3}NE. This support for batch parallelism is limited in CPUs, like the Intel i7-6700K, which affects both the latency and the throughput. TrueNorth by IBM~\cite{merolla2014million} is a custom neuromorphic ASIC, which demonstrates the potential of SNN hardware when optimized for low power consumption. Its lack of malleability, however, inhibits the scalability and generality, as network models have to be specifically designed and trained to be compatible with the static architecture. Similarly, Intel's Loihi chip imposes restrictions on the number of inputs of each neuron core, which requires a special adaptation and mapping of a network structure to the chip's architecture~\cite{davies2018loihi}.

As expected from a generic framework, E\textsuperscript{3}NE can deploy a wide variety of convolutional SNN. That includes small models like LeNet, which has a relatively low number of 61k parameters. We used 3-bit weights and four time steps for activations in this experiment. That allows all data to be stored in on-chip memory, avoiding external DRAM access. That contributes to the comparably low energy consumption. Another power-saving factor is the small amount of logic resources required, even with four parallel convolution modules being used. In fact, a large amount of energy could be saved by deploying the accelerator on a much smaller FPGA device. When synthesized for an XCKU3P, the smallest part with the Xilinx Ultrascale+ architecture, a mere 1.2 W of power dissipation were reported. The inference of LeNet has accuracy of 99.09\%, while processing 3400 images per second. On the other end of the spectrum, two larger networks AlexNet~\cite{krizhevsky2012imagenet} and VGG-11~\cite{simonyan2014very} were executed. The CIFAR datasets come closer to real-world image data, as they picture photographs of objects in their natural environment. That makes them considerably more challenging to classify than MNIST's handwritten digits. Moreover, the CIFAR-100 dataset contains objects of 100 different classes. Understandably, the SNN models are significantly larger containing more than 23 million parameters each. Due to the increased complexity of both SNN model and dataset, six time steps and 6-bit weights are used to reach an acceptable accuracy. The quantity of parameters and layers influences the latency, which falls into the range of milliseconds for both AlexNet and VGG. Eight parallel convolution modules have been used to mitigate the relatively high runtime. To no surprise, this and the need for external memory accesses result in higher power and logic resources. Despite the large models, less hardware resources are used compared with~\cite{ju2020fpga} and~\cite{fang2020encoding}, displaying the area efficiency and scalability of E\textsuperscript{3}NE.

\section{Conclusion} \label{sec: conclusion}
In this paper, we presented E\textsuperscript{3}NE, an end-to-end framework for the inference of spiking neural networks with emerging neural encoding. It relies on an RTL hardware library, which contains area efficient processing modules for common SNN operations. Their capability to be dynamically reconfigured is exploited by the compiler, whose objective is an optimal utilization of the instantiated hardware for the sake of efficiency and performance. Both of those goals are also addressed by the emerging neural encoding scheme. We describe the technique to generate spike trains from real-valued input samples using dynamic quantization. The execution is controlled by a sequence of instructions, which is generated in accordance with the layers' loop hierarchies. Through loop interchange and unrolling, we found a balance between processing speed and memory requirements. The reordering of instructions enabled parallelism on a macro-level, as two different segments of the logic were operating concurrently.

A variety of experiments demonstrated the impact of optimizations and trade-offs found in E\textsuperscript{3}NE. Besides the importance of parallelism for hardware utilization, we observed how the spike train length affects the performance and accuracy. When compared with previous FPGA-based SNN accelerators, E\textsuperscript{3}NE outperforms with respect to all metrics. Especially the superiority in terms of power and hardware resources allowed us to deploy larger SNN models on the FPGA accelerator.

\section*{Acknowledgments}
This work was supported by the Singapore Government’s Research, Innovation and Enterprise 2020 Plan (Advanced Manufacturing and Engineering domain) under Grant A1687b0033.

\bibliographystyle{IEEEtran}
\bibliography{references}

\begin{IEEEbiography}[{\includegraphics[width=1in,height=1.25in,clip,keepaspectratio]{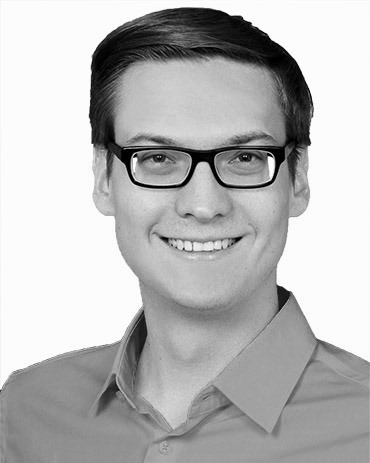}}]{Daniel Gerlinghoff}
was awarded a bachelor's degree in electrical engineering by the Leipzig University of Applied Sciences, Germany in 2017. He carried out internships as software developer and digital circuit designer, before continuing his studies towards a master's degree in integrated circuit design at Nanyang Technological University, Singapore. As part of his dissertation, he implemented a neural network inference accelerator on FPGA, which was tightly constrained by power and logic resources. After his graduation in 2020, he continues research on FPGA-based machine learning and heterogeneous computing as a research engineer at Institute of High Performance Computing, Agency for Science Technology and Research in Singapore.
\end{IEEEbiography}

\begin{IEEEbiography}[{\includegraphics[width=1in,height=1.25in,clip,keepaspectratio]{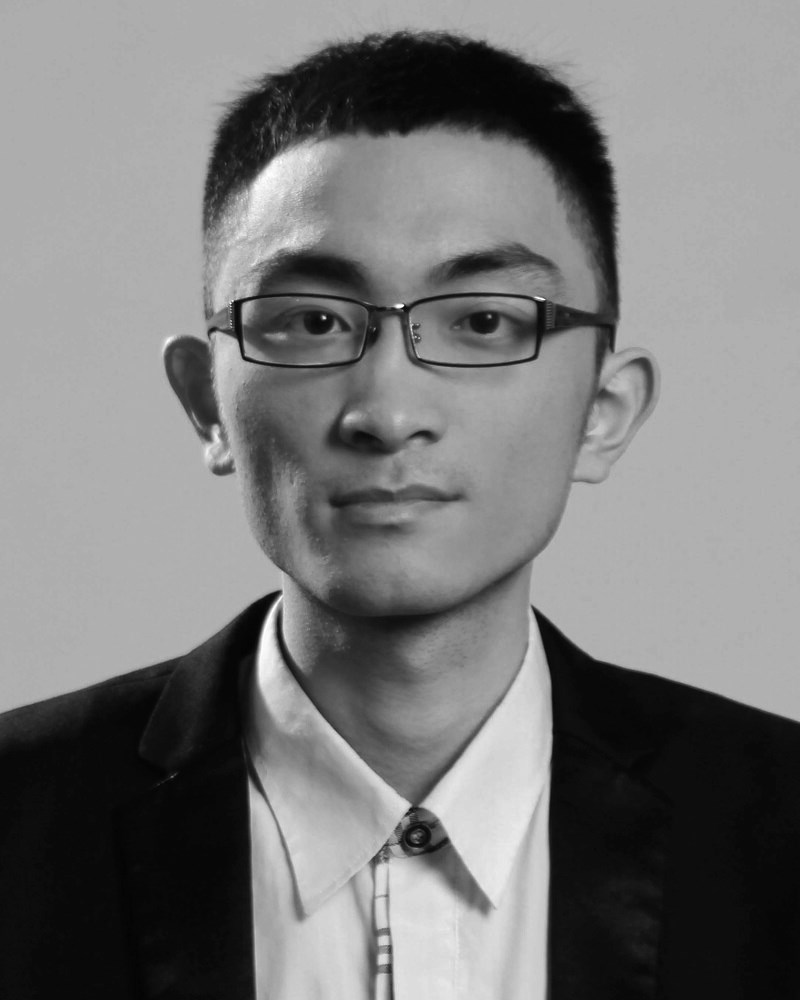}}]{Zhehui Wang}
received B.S. degree in Electrical Engineering from Fudan University, China, in 2010, and Ph.D. degree in Electronic and Computer Engineering from Hong Kong University of Science and Technology, Hong Kong, in 2017. He is currently a scientist of Agency for Science, Technology and Research (A*STAR), Singapore. His research interests include Machine learning, model compression, and neuromorphic chip.
\end{IEEEbiography}

\begin{IEEEbiography}[{\includegraphics[width=1in,height=1.25in,clip,keepaspectratio]{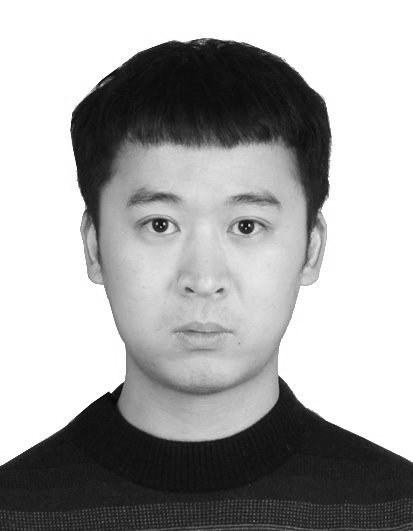}}]{Xiaozhe Gu}
received the Ph.D. degree in computer science and engineering from Nanyang Technological University, Singapore, in 2018. He is currently a Post-Doctoral Researcher with the Future Network of Intelligence Institute, the Chinese University of Hong Kong, Shenzhen. His research interests include real-time system, scheduling algorithm and machine learning.
\end{IEEEbiography}


\begin{IEEEbiography}[{\includegraphics[width=1in,height=1.25in,clip,keepaspectratio]{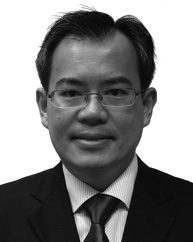}}]{Rick Siow Mong Goh}
received the Ph.D. degree in electrical and computer engineering from the National University of Singapore, Singapore.

He is the Director of the Computing \& Intelligence (CI) Department, Institute of High Performance Computing, Agency for Science, Technology and Research, Singapore, where he leads a team of over 80 scientists in performing world-leading scientific research, developing technology to commercialization, and engaging and collaborating with industry. His current research interests include artificial intelligence, high-performance computing, block chain, and federated learning.
\end{IEEEbiography}

\begin{IEEEbiography}[{\includegraphics[width=1in,height=1.25in,clip,keepaspectratio]{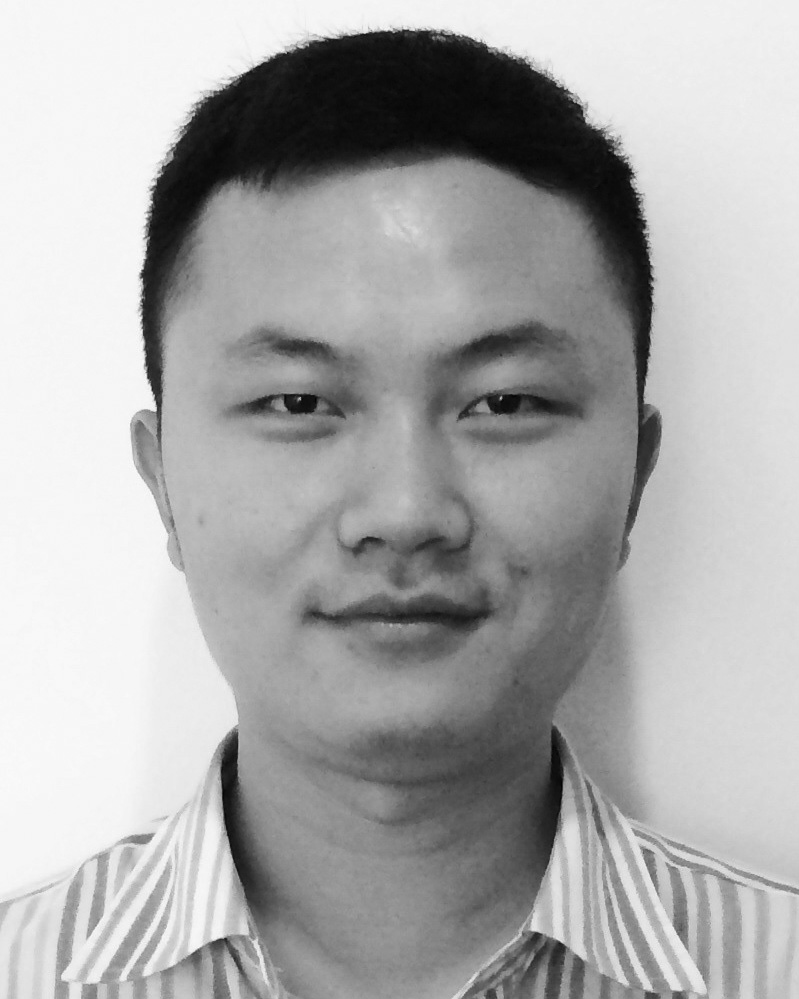}}]{Tao Luo}
received the bachelor’s degree from the Harbin Institute of Technology, Harbin, China, in 2010, the master’s degree from the University of Electronic Science and Technology of China, Chengdu, China, in 2013, and the Ph.D. degree from the School of Computer Science and Engineering, Nanyang Technological University, Singapore, in 2017.

He is currently a Research Scientist with the Institute of High Performance Computing, Agency for Science, Technology and Research, Singapore. His current research interests include high-performance computing with emerging device, artificial intelligence, reconfigurable computing system, and cyber security.
\end{IEEEbiography}

\vfill

\end{document}